\documentclass[11pt,a4paper]{article}
\usepackage[hyperref]{emnlp2020}
\usepackage{times}
\usepackage{latexsym}

\usepackage{microtype}
\usepackage{graphics}
\usepackage{graphicx}

\aclfinalcopy % Uncomment this line for the final submission
\title{Corpora Compared: The Case of the Swedish Gigaword \& Wikipedia Corpora}

\author{Tosin P. Adewumi\thanks{Corresponding author | Presented at the Eighth Swedish Language Technology Conference (SLTC)} \\
  \\
  \\
  \texttt{} \\\And
  Foteini Liwicki \\ \\
  Machine Learning group, EISLAB %/ Address line 1 
  \\
  Luleå University of Technology, Sweden. % / Address line 2 \\Affiliation / Address line 3
  \\
  \texttt{firstname.lastname@ltu.se} \\\And
  Marcus Liwicki \\
  \\
  \\
  \texttt{}
  \\}

\date{}

\begin{document}
\maketitle
\begin{abstract}
In this work, we show that the difference in performance of embeddings from differently sourced data for a given language can be due to other factors besides data size.
Natural language processing (NLP) tasks usually perform better with embeddings from bigger corpora.
However, broadness of covered domain and noise can play important roles.
We evaluate embeddings based on two Swedish corpora: The Gigaword and Wikipedia, in analogy (intrinsic) tests and discover that the embeddings from the Wikipedia corpus generally outperform those from the Gigaword corpus, which is a bigger corpus.
Downstream tests will be required to have a definite evaluation.
\end{abstract}

\section{Introduction}
It is generally observed that more data bring about better performance in Machine Learning (ML) tasks \cite{adewumi2019conversational, stevens2020deep}.
What may not be very clear is the behaviour of variance of homogeneity in datasets.
It is always better to have a balanced or broad-based dataset or avoid an overly-represented topic within a dataset \cite{stevens2020deep}.
Furthermore, noise (or contamination) in data
can reduce performance \cite{hagan1997neural}.
However, not all noise is bad.
Indeed, noise may be helpful \cite{stevens2020deep}.

In this work, we compare embeddings (in analogy test) from two Swedish corpora: The Gigaword and Wikipedia.
The Gigaword corpus by \citet{rodven2016swedish} contains data from different genre, covering about 7 decades since the 1950s.
Meanwhile the Wikipedia is a collection of articles on many, various subjects \cite{svwiki}.

Word similarity or analogy tests, despite their weaknesses, have been shown to reveal somewhat meaningful relationships among words in embeddings, given the relationship among words in context \cite{mikolov2013efficient, pennington2014glove}.
It is misleading to assume such intrinsic tests are sufficient in themselves, just as it is misleading to assume one particular extrinsic (downstream) test is sufficient to generalise the performance of embeddings on all NLP tasks \cite{gatt2018survey, faruqui2016problems, adewumi2020word2vec}.

The research question being addressed in this work is: does bigger corpus size automatically mean better performance for differently-sourced Swedish corpora?
The contribution this work brings is the insight into the differences in the performance of the Swedish embeddings of the Gigaword and Wikipedia corpora, despite the over 40\% additional size of the Gigaword corpus.
Furthermore, this work will, possibly, enable researchers seek out ways to improve the Gigaword corpus, and indeed similar corpora, if NLP downstream tasks confirm the relative better performance of embeddings from the Wikipedia corpus.
The following sections include related work, methodology, results \& discussion and conclusion.

\section{Related Work}
\citet{rodven2016swedish} created the Swedish corpus with at least one billion words.
It covers fiction, government, news, science and social media from the 1950s.
The sentences of the first six lines of the content of this Gigaword corpus are:
\begin{quote}
1 knippa dill\\
patrik andersson\\
TV : Danska Sidse Babett Knudsen har prisats på tv-festivalen i Monte Carlo för rollen\\ i dramaserien Borgen .\\
Hon sköts med ett skott i huvudet , men tog sig fram till porten och ringde på .\\
I början av juni tog hon examen från den tvååriga YH-utbildning , som hon flyttade upp till huvudstaden för att gå .\\
Det blev kaos , folk sprang fram för att hjälpa , någon började filma ...
\end{quote}

The content of the Wikipedia corpus is a community effort, which began some years ago, and is edited continually.
It covers far-reaching topics, including those of the Swedish Gigaword corpus, and in addition, entertainment, art, politics and more.
The sentences of the first seven lines of the content of the pre-processed version of the Wikipedia corpus are given below.
It would be observed that it contains a bit of English words and the pre-processing script affected non-ascii characters.
However, these issues were not serious enough to adversely affect the models generated, in this case, as the embedding system seems fairly robust to handle such noise.
\begin{quote}amager r en dansk i resund ns norra och v stra delar tillh r k penhamn medan vriga delar upptas av t rnby kommun och drag rs kommun amager har en yta p nine six two nine km och befolkningen uppg r till one nine six zero four seven personer one one two zero one eight en stor del av bebyggelsen har f rortspr gel men ven tskilliga innerstadskvarter finns i k penhamn samt i drag r p den stra delen av n finns kastrups flygplats amager r delvis en konstgjord delvis en naturlig s dan n r mycket l g och vissa delar ligger under havsytan framf r allt det genom f rd mning.
\end{quote}

\citet{adewumi2020exploring} created the Swedish analogy test set, which is similar to the Google analogy test set by  \citet{mikolov2013efficient}.
This was because there was no existing analogy test set to evaluate Swedish embeddings \cite{fallgren2016towards, precenth2019word}.
The analogy set has two main sections and their corresponding subsections: the semantic \& syntactic sections.
Two native speakers proof-read the analogy set for any possible issues (with percentage agreement of 98.93\% between them), after valuable comments from the reviewers of this paper.
It is noteworthy that some words can have two or more possible related words.
For example, based on the dictionary, the Swedish word \textit{man} can be related to \textit{kvinna} and \textit{dam} in very similar ways.
Four examples from the \textit{gram2-opposite} sub-section of the syntactic section are:
\begin{quote}
medveten omedveten lycklig olycklig\\
medveten omedveten artig oartig\\
medveten omedveten härlig ohärlig\\
medveten omedveten bekväm obekväm
\end{quote}

\citet{faruqui2016problems} correctly suggest there are problems with word similarity tasks for intrinsic evaluation of embeddings.
One of the problems is overfitting, which large datasets (like the analogy set in this work) tend to alleviate \cite{stevens2020deep}.
In order to have a definite evaluation of embeddings, it's important to conduct experiments on relevant downstream tasks \cite{faruqui2016problems, faruqui2014improving, lu2015deep, gatt2018survey}.

\section{Methodology}
Table 1 gives the meta-data of the two corpora used.
The Gigaword corpus was generated as described by \citet{rodven2016swedish} while the Wikipedia corpus was pre-processed using the recommended script by \cite{grave2018learning}.
This script returned all text as lowercase and does not always retain non-asci characters.
This created noise in the corpus, which may not necessarily be harmful, as it has been shown in a recent work that diacritics can adversely affect performance of embeddings unlike their normalized versions \cite{adewumi2020challenge}.
A portion of the pre-processed text (given in the previous section) was also tested for coherence on Google Translate and the English translation returned was meaningful, despite the noise.
Hence, the noise issue was not serious enough to adversely affect the models generated in this case, as the embedding system seems fairly robust to handle such noise.

\begin{table}[h]
\centering
%\resizebox{\columnwidth}{!}{%
\begin{tabular}{c|c|c}
\textbf{Meta-data} & \textbf{Gigaword} & \textbf{Wikipedia}
\\
\hline
Size & 5.9G & 4.2G
\\
\hline
Tokens & 1.08B & 767M
\\
\hline
Vocabulary & 1.91M & 1.21M
\\
\hline
Year & 2016 & 2019
\\
\hline
\end{tabular}
%}
\label{table:meta}
\caption{Meta-data for both Swedish Corpora}
\end{table}

The authors made use of the fastText C++ library (with default hyper-parameters, except where mentioned) by \citet{grave2018learning} to generate 8 word2vec models and 8 subword models from each corpus, based on the optimal hyper-parameter combinations demonstrated by \citet{adewumi2020word2vec}.
Each model was intrinsically evaluated using the new Swedish analogy test set by \citet{adewumi2020exploring} in a Python-gensim program \cite{rehurek_lrec}.
%Regardless of any possible weakness in the analogy set, the fact that it was the same set used as a benchmark for the two corpora creates a level field for evaluation.
The hyper-parameters tuned are window size (4 \& 8), neural network architecture (skipgram \& continuous bag of words(CBoW)) and loss (heirarchical softmax and negative sampling).
The subword models used lower \& upper character n-gram values of 3 \& 6, respectively.

Although each model in the first set of experiments, with default (starting) learning rate (LR) of 0.05, was run twice and average analogy score calculated, it would have been more adequate to calculate averages over more runs per model and conduct statistical significance tests.
Nonetheless, the statistical significance tests can be conducted for the downstream tasks, which usually are the key tests for the performance of these embeddings.
It should also be noted that deviation from the mean of each model performance for their corresponding two runs is minimal.
Due to the observation of one model (of Gigaword-CBoW-hierarchical softmax) failing (with \textit{Encountered NaN} error) when using the default LR of 0.05, another set of experiments with the LR of 0.01 was conducted but with single run per model, due to time constraint.

\section{Results \& Discussion}
Table 2 gives mean analogy scores for LR 0.05 of embeddings for the two corpora and table 3 for LR of 0.01.
It will be observed that the skipgram-negative sampling combination for both corpora for word2vec and subword models performed best in both tables, except one in table 3, confirming what is known from previous research \cite{mikolov2013efficient, adewumi2020word2vec, adewumi2020exploring}.
From table 2, the highest score is 60.38\%, belonging to the word2vec embedding of the Wikipedia corpus.
The lowest score is 2.59\%, belonging to the CBoW-hierarchical softmax, subword embedding of the Gigaword corpus.
The highest score in table 3 also belongs to the Wikipedia word2vec model.
Among the 8 embeddings in the word2vec category in table 2, there are 6 Wikipedia embeddings with greater scores than the Gigaword while among the subword, there are 5 Wikipedia embeddings with greater scores.
Nearest neighbour qualitative evaluation of the embeddings for a randomly selected word is given in table 4.
%Kilgarriff:2001 mention to do away with hypothesis testing with regards to corpora comparison

%Chi-square test outperforms other types of tests like Spearman correlation coefficient & variants of cross-entropy measures

\begin{table}[h]
\centering
\resizebox{\columnwidth}{!}{%
\begin{tabular}{c|c|c|c|c|c|c|c|c}
\textbf{} &
\multicolumn{4}{c}{\textbf{Skipgram (s1)}} & \multicolumn{4}{|c|}{\textbf{CBoW (s0)}}
\\
\textbf{} &
\multicolumn{2}{c}{{\textbf{H. S. (h1)}}} &
\multicolumn{2}{|c}{{\textbf{N. S. (h0)}}} &
\multicolumn{2}{|c}{{\textbf{H. S. (h1)}}} &
\multicolumn{2}{|c|}{{\textbf{N. S. (h0)}}}
\\
\hline
\textbf{window (w)} &
{\textbf{4}} & {\textbf{8}} & {\textbf{4}} &
{\textbf{8}} & {\textbf{4}} & {\textbf{8}} & {\textbf{4}} & {\textbf{8}}
\\
\hline
{\textbf{Word2Vec \%}} &
\multicolumn{8}{c}{}
\\
\hline
{Wikipedia} &
{47.02} & {44.09} & {\textbf{60.38}} & {60.38} & {29.09} & {30.09} & {54.39} & {56.81}
\\
\hline
{Gigaword} &
{40.26} & {44.23} & {\textbf{55.79}} & {55.21} & {26.23} & {27.82} & {55.2} & {55.81}
\\
\hline
{\textbf{Subword \%}} &
\multicolumn{8}{c}{}
\\
\hline
{Wikipedia} &
{46.65} & {45.8} & {\textbf{56.51}} & {56.36} & {28.07} & {24.95} & {38.26} & {35.92}
\\
\hline
{Gigaword} &
{41.37} & {44.7} & {\textbf{58.31}} & {56.28} & {2.59} & {-} & {46.81} & {46.39}
\\
\hline
\end{tabular}
}
\label{table:subwords}
\caption{Mean Analogy Scores for Swedish Gigaword \& Wikipedia Corpora with LR=0.05}
\end{table}

\begin{table}[h]
\centering
\resizebox{\columnwidth}{!}{%
\begin{tabular}{c|c|c|c|c|c|c|c|c}
\textbf{} &
\multicolumn{4}{c}{\textbf{Skipgram (s1)}} & \multicolumn{4}{|c|}{\textbf{CBoW (s0)}}
\\
\textbf{} &
\multicolumn{2}{c}{{\textbf{H. S. (h1)}}} &
\multicolumn{2}{|c}{{\textbf{N. S. (h0)}}} &
\multicolumn{2}{|c}{{\textbf{H. S. (h1)}}} &
\multicolumn{2}{|c|}{{\textbf{N. S. (h0)}}}
\\
\hline
\textbf{window (w)} &
{\textbf{4}} & {\textbf{8}} & {\textbf{4}} &
{\textbf{8}} & {\textbf{4}} & {\textbf{8}} & {\textbf{4}} & {\textbf{8}}
\\
\hline
{\textbf{Word2Vec \%}} &
\multicolumn{8}{c}{}
\\
\hline
{Wikipedia} &
{48.92} & {49.01} & 51.71 & \textbf{53.48} & {32.36} & {33.92} & {47.05} & {49.76}
\\
\hline
{Gigaword} &
{39.12} & {43.06} & 48.32 & \textbf{49.96} & {28.89} & {31.19} & {44.91} & {48.02}
\\
\hline
{\textbf{Subword \%}} &
\multicolumn{8}{c}{}
\\
\hline
{Wikipedia} &
{45.16} & \textbf{46.82} & 35.91 & {43.26} & {22.36} & {21.1} & {14.31} & {14.45}
\\
\hline
{Gigaword} &
{39.13} & {43.65} & 45.51 & \textbf{49.1} & {31.67} & {35.07} & {28.34} & {28.38}
\\
\hline
\end{tabular}
}
\label{table:lr01}
\caption{Analogy Scores for Swedish Gigaword \& Wikipedia Corpora with LR=0.01}
\end{table}

\begin{figure}[!htb]
   \begin{minipage}{.5\textwidth}
     \centering
     \includegraphics[width=1\linewidth]{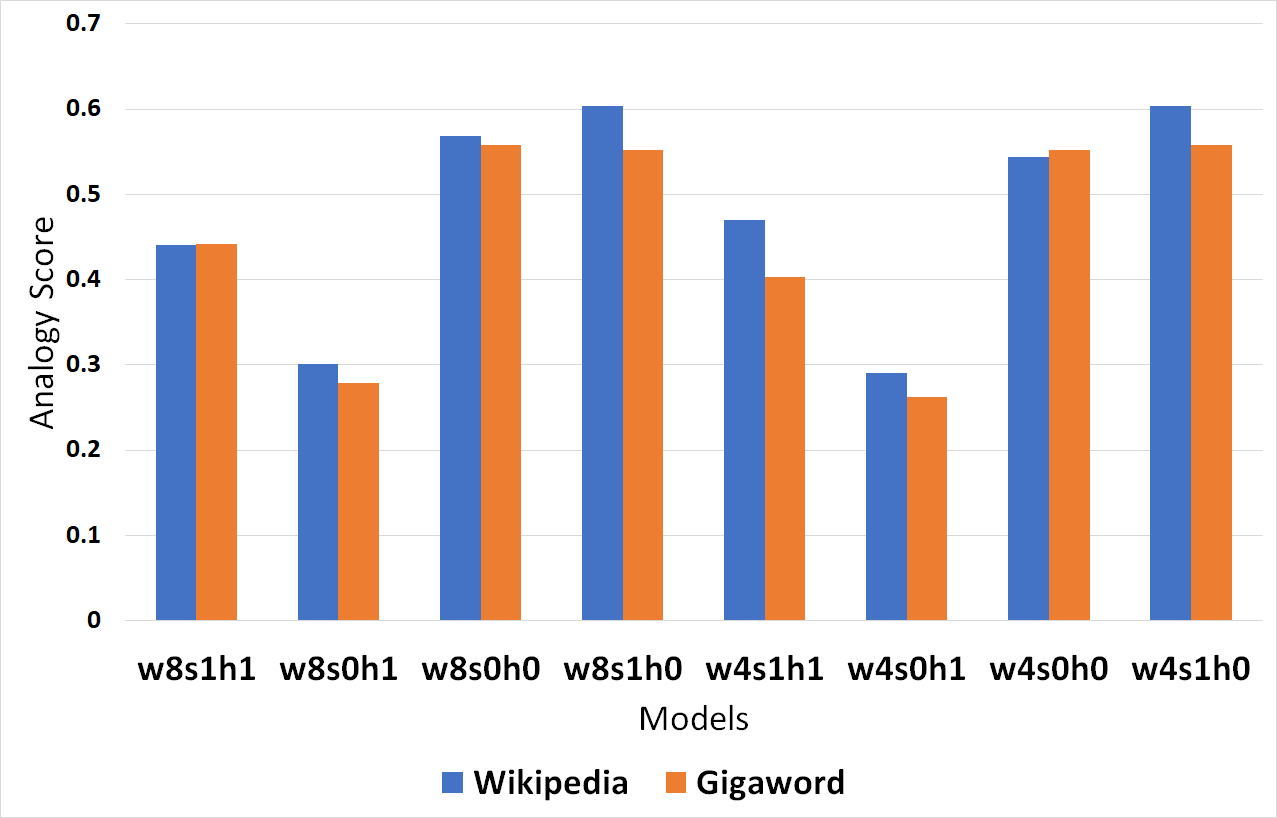}
     \caption{Word2Vec Mean  Scores, LR:0.05}\label{Fig:engNER}
   \end{minipage}\hfill
   \begin{minipage}{.5\textwidth}
     \centering
     \includegraphics[width=1\linewidth]{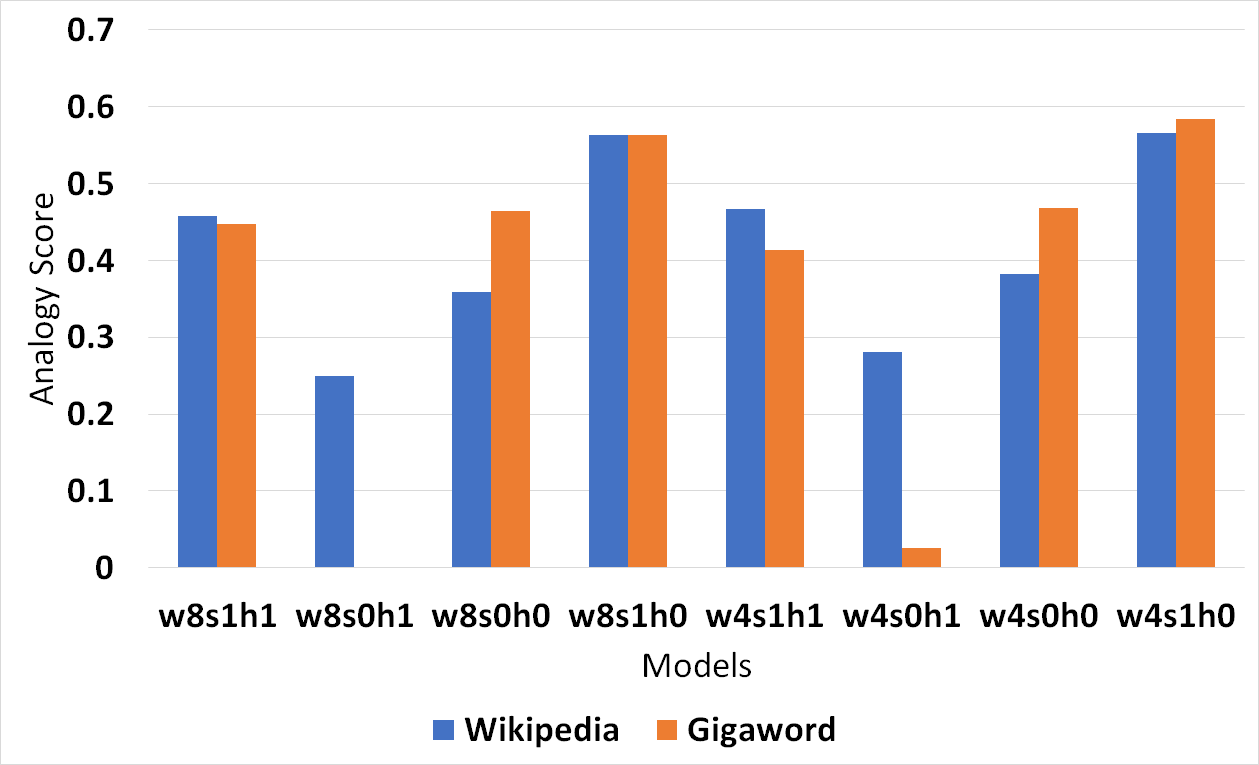}
     \caption{Subword Mean Scores, LR:0.05}\label{Fig:sweNER}
   \end{minipage}
\end{figure}

\begin{table}[hbt!]
\centering
\resizebox{\columnwidth}{!}{%
\begin{tabular}{c|c}
\textbf{\footnotesize{Nearest Neighbor}} & \textbf{\footnotesize{Result}} \\
\hline
\footnotesize{Wiki: syster} & \footnotesize{systerdotter (0.8521), systern (0.8359), ..} \\
\hline
\footnotesize{Gigaword: syster} & \footnotesize{systerdotter (0.8321), systerdottern (0.8021), ..}\\
\hline
\end{tabular}
}
\label{quality}
\caption{Example qualitative assessment of Swedish subword w4s1h0 models}
\end{table}

We hypothesize that the general performance difference observed between the embeddings of the two corpora may be due to a) the advantage of wider domain coverage (or corpus balance in topics) of the Wikipedia corpus - which is the most plausible reason, b) the small noise in the Wikipedia corpus or c) the combination of both earlier reasons.

Since it's preferable to have more than one criterion for the difference between the two corpora, future work will focus, particularly, downstream tasks to confirm this \cite{faruqui2016problems, gatt2018survey}.
Implementation without using the pre-processing script by \citep{grave2018learning} on the original Wikipedia corpus will also be attempted.

\section{Conclusion}
This work has shown that better performance results from differently sourced corpora of the same language can be based on reasons besides larger data size.
Simply relying on larger corpus size for performance may be disappointing.
The Wikipedia corpus showed better performance in analogy tests compared to the Gigaword corpus.
Broad coverage of topics in a corpus seems important for better embeddings and noise, though generally harmful, may be helpful in certain instances.
Future work will include other tests and downstream tasks for confirmation.

\section{Acknowledgement}
The authors wish to thank the anonymous reviewers for their valuable contributions and the very useful inputs from Carl Borngrund and Karl Ekström, who proof-read the analogy set.
The work on this project is partially funded by Vinnova under the project number 2019-02996 ”Sprkmodeller fr svenska myndigheter”.

\bibliographystyle{acl_natbib}
\bibliography{emnlp2020}

\end{document}